\documentclass{article}

 \usepackage{neurips_2025}
\usepackage{makecell}
\usepackage[utf8]{inputenc} 
\usepackage[T1]{fontenc}    
\usepackage{hyperref}       
\usepackage{url}            
\usepackage{booktabs}       
\usepackage{amsfonts}       
\usepackage{nicefrac}       
\usepackage{microtype}      
\usepackage{xcolor}         
\usepackage{multirow}
\usepackage{graphicx}
\usepackage{booktabs}
\usepackage{algorithm}
\usepackage{algorithmic}
\usepackage[accsupp]{axessibility}  
\usepackage{amssymb}   
\usepackage{pifont}    
\usepackage[table]{xcolor}
\usepackage{natbib}
\title{M$^3$-ACE: Rectifying Visual Perception in Multimodal Math Reasoning via Multi-Agentic Context Engineering}

\author{%
  Peijin Xie \thanks{Equal contribution.}\\
  ITNLP Lab\\
  Harbin Institute of Technology\\
  \texttt{xpj@stu.hit.edu.cn} \\
  \And
  Zhen Xu \footnotemark[1] \\
  Platform and Content Group \\
  Tencent \\
  \texttt{zenxu@tencent.com} \\
  \AND
  Bingquan Liu \thanks{Corresponding author.} \\
  ITNLP Lab\\
  Harbin Institute of Technology\\
  \texttt{ bqliu@hit.edu.cn} \\
  \And
  Baoxun Wang \footnotemark[2]\\
  Platform and Content Group \\
  Tencent \\
  \texttt{asulewang@tencent.com} \\
}

\begin{document}

\maketitle

\begin{abstract}

Multimodal large language models have recently shown promising progress in visual mathematical reasoning. 
However, their performance is often limited by a critical yet underexplored bottleneck: inaccurate visual perception. 
Through systematic analysis, we find that the most failures originate from incorrect or incomplete \emph{visual evidence} extraction rather than deficiencies in reasoning capability. 
Moreover, models tend to remain overly confident in their initial perceptions, making standard strategies such as prompt engineering, multi-round self-reflection, or posterior guidance insufficient to reliably correct errors.

To address this limitation, we propose \textbf{M$^3$-ACE}, a multi-agentic context engineering framework designed to rectify visual perception in multimodal math reasoning. 
Instead of directly aggregating final answers, our approach decouples perception and reasoning by dynamically maintaining a shared context centered on visual evidence lists. 
Multiple agents collaboratively contribute complementary observations, enabling the system to expose inconsistencies and recover missing perceptual information. 
To support stable multi-turn collaboration, we further introduce two lightweight tools: a \emph{Summary Tool} that organizes evidence from different agents into consistent, complementary, and conflicting components, and a \emph{Refine Tool} that filters unreliable samples and guides iterative correction.

Extensive experiments demonstrate that M$^3$-ACE substantially improves visual mathematical reasoning performance across multiple benchmarks. 
Our method establishes new state-of-the-art results 89.1 on the \textbf{MathVision} benchmark and achieves consistent improvements on other related datasets, including \textbf{MathVista} and \textbf{MathVerse}. 
These results highlight the importance of perception-centric multi-agent collaboration for advancing multimodal reasoning systems.

\end{abstract}

\section{Introduction}
\label{sec:intro}

Visual mathematical reasoning has emerged as a critical benchmark for evaluating the integrated capabilities of modern vision-language models (VLMs), for example, requiring accurate visual perception\citep{vlmprobing}, symbolic abstraction\citep{xu2025diagrams}, and multi-step logical reasoning\citep{liu2025vlmfo1}. 
Recent advances in large-scale multimodal models have significantly improved reasoning performance on visual math tasks, particularly under chain-of-thought\citep{jia2025fast} or tool-augmented prompting paradigms\citep{yin2025toolvqa}. 
However, despite these gains, model performance remains fragile when confronted with subtle perceptual ambiguities, noisy visual evidence\citep{zhang2025viper}, or complex diagrammatic representations\citep{liu2025perception, li2025chain}.

In parallel, context engineering has recently gained attention as an alternative to parameter-level optimization \citep{chen2025prompt}, demonstrating that carefully structured prompts \citep{xu2025llavacot}, intermediate representations \citep{liu2025figr}, and interaction protocols \citep{lin2025mind} can substantially enhance model performance without additional training. This paradigm shift suggests that failures in visual reasoning may not solely originate from insufficient reasoning capacity, but rather from how perceptual information is extracted, structured \citep{chung2025mllm,wu2026vision}, and fed into the reasoning process \citep{yang2025strategic}.

\begin{figure}[htbp]
  \centering
  \includegraphics[height=6.5cm]{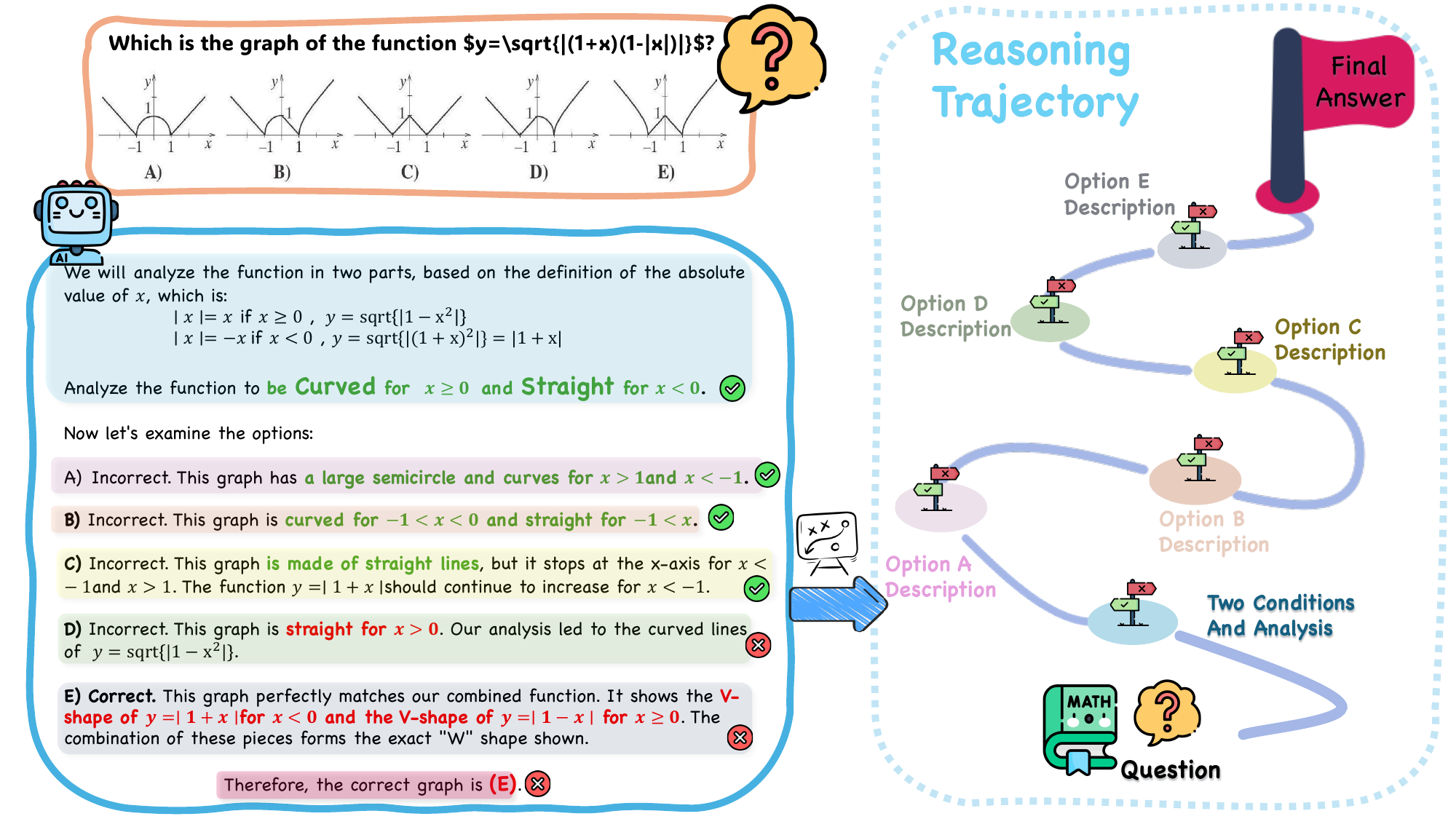}
  \caption{Visual Evidence as the Bottleneck — Decoupling Perception and Reasoning.
  }
  \label{fig:introduction}
\end{figure}
In this work, we revisit visual mathematical reasoning from the perspective of Visual Evidence (VE), i.e., the structured perceptual facts extracted from images that serve as the foundation for downstream reasoning. 
We argue that reasoning and perception should be explicitly decoupled: while reasoning traces generated by state-of-the-art (SOTA) models are often logically sound, failures frequently arise from incorrect or incomplete visual evidence.


Empirically, perception has become the dominant bottleneck in solving visual math problems \citep{endo2026downscaling,chen2026drawedumath,wang2025perception}. Prior open-source approaches largely relied on reinforcement learning or task-specific fine-tuning to mitigate this issue \citep{mars2025challenge,vllm2026survey}. In contrast, under the era of powerful SOTA foundation models, we demonstrate that context engineering alone---without additional training ---can effectively address perception errors by restructuring how visual evidence is elicited, verified, and refined. This distinguishes our approach from prior works that focus on improving reasoning ability through model updates rather than improving perceptual reliability through interaction design \citep{maxim2025prompt,cotvla2025cvpr}

We first conduct a systematic empirical study to verify that visual evidence extraction is indeed the primary source of errors in state-of-the-art visual mathematical reasoning models. 
By decoupling the reasoning process into visual evidence extraction, trajectory planning, and final reasoning using Gemini-2.5 Pro and GPT-5, we perform a detailed bad-case analysis alongside a two-round mutual correction experiment.

Our findings reveal a strong correlation between answer correctness and VE correctness. 
Specifically:
(1) For all incorrect answers, the error can be traced back to incorrect visual evidence, while the corresponding reasoning traces remain logically valid.
(2) For correct answers, both the extracted visual evidence and reasoning traces are consistently correct.

Furthermore, we observe a striking asymmetry: providing the model with corrected visual evidence can successfully rectify wrong answers, whereas supplying the correct final answer does not enable the model to recover the correct visual evidence. 
These results indicate that models fundamentally struggle with visual evidence correction, and that correct VE is a necessary precondition for correct answers, rather than a byproduct of successful reasoning.

We further investigate whether a single model can autonomously correct its visual evidence errors within an isolated single-model setting. 
On the same benchmark, we explore several commonly adopted self-correction strategies, including:
(1) prompt engineering in a single-round question-answering setup;
(2) multi-round self-reflection with reflection on the final answer;
(3) multi-round interactions with explicit reflection on visual evidence;
(4) explicitly informing the model that its visual evidence is incorrect and asking for revision.

Across all settings, we find that single-model self-correction yields only marginal improvements and fails to reliably fix VE errors. 
Notably, prompt engineering and multi-round self-reflection without external guidance may even destabilize correct predictions. These results suggest that once incorrect visual evidence is formed, the model exhibits strong confirmation bias and lacks the capability to effectively re-perceive the image, even when guided by reflective prompts. These findings motivate approaches beyond isolated single-model reflection.

Motivated by these observations, we propose M3-Agent, a multi-agent framework designed to progressively address the visual evidence bottleneck. 
Our system is built upon three core design principles: 
(i) \textbf{Decoupling Principle}, which explicitly separates the visual evidence list from the final answer to directly target the bottleneck; 
(ii) \textbf{Complementary Information Principle}, which introduces heterogeneous assistant agents to provide diverse and potentially conflicting visual evidence, thereby reducing confirmation bias in self-reflection; 
and (iii) \textbf{Filtering Principle}, which selectively focuses refinement efforts on difficult or highly disputed samples while filtering out easy or high-consensus cases early.

Instead of relying on a single model's self-reflection, our approach introduces structured multi-agent interactions over a shared visual evidence context, enabling cross-validation through consistent, complementary, and conflicting evidence categorization, disagreement resolution via lightweight summary and refinement tools, and iterative refinement of perceptual facts until convergence.

Our main contributions are threefold:

\begin{enumerate}
    \item We identify and empirically validate visual evidence extraction as the primary bottleneck in state-of-the-art visual mathematical reasoning models, showing that errors primarily originate from incorrect visual evidence rather than flawed reasoning.
    \item We demonstrate that single-model self-correction is insufficient for resolving VE errors—prompt engineering and multi-round self-reflection without external guidance fail to improve visual evidence accuracy and may destabilize correct predictions.
    \item We propose M3-Agent, a novel multi-agent context engineering framework that significantly improves visual evidence accuracy through structured cross-validation and iterative refinement, consequently enhancing final answer correctness without additional model training.
\end{enumerate}

\section{Related Work}
\label{related work}

\subsection{Visual Math Reasoning}
The evaluation of Large Multimodal Models (LMMs) in visual mathematical reasoning has been significantly advanced by three pivotal benchmarks ~\citep{xue2025mmrc,wu2025alignmmbench} that collectively address different dimensions of the problem. 
MathVista~\citep{mathvista} serves as the first comprehensive foundation benchmark, integrating 28 existing datasets with three newly developed subsets (IQTest, FunctionQA, and PaperQA) to provide 6,141 samples spanning seven distinct reasoning types including algebraic, geometric, and statistical reasoning. 
Building upon this foundation, MathVerse~\citep{mathverse} introduces an innovative ``Visual Ablation'' design that transforms 2,612 original problems into over 15,000 test samples across six versions ranging from text-dominant to vision-only configurations, enabling precise diagnosis of whether models genuinely utilize visual information or merely rely on textual redundancy \citep{yang2025visualredundancy,lv2025llmcplus}. 
Representing the upper bound of difficulty, MathVision~\citep{mathvision} focuses on competition-level challenges by sourcing 3,040 high-quality problems from prestigious mathematics competitions such as AMC and AIME, covering 16 mathematical disciplines across five difficulty levels \citep{xu2025mars2}.

Recent models have demonstrated remarkable progress on these benchmarks: the Qwen3.5 series achieves 85.7\% on MathVista and 78.9\% on MathVision, substantially surpassing human baselines of 60.3\% and 68.82\% respectively. 
However, several critical challenges persist \citep{li2025perception,zhu2025deepthinking}. 
First, the significant performance gap between MathVista (85.7\%) and MathVerse (63.5\%) for the same model reveals ``modality laziness'' \citep{yan2025dataquality}, where models tend to rely heavily on textual cues while insufficiently leveraging visual information. 
Second, current models struggle with fine-grained visual perception \citep{yan2025visuriddles,wang2025finegrained,wang2025cof}, particularly in parsing geometric diagrams with precise spatial relationships and symbolic annotations \citep{ma2025deepperception,jiang2025finegrained}. 
Third, the substantial gap on competition-level problems (GPT-4o achieving only 30.39\% on MathVision) underscores difficulties in orchestrating complex multi-step reasoning chains \citep{li2025vocot,plaat2025multistep,thawakar2025llamavo1} that integrate visual understanding with mathematical formalization \citep{su2025thinkingimages,zhan2025understand}. 
These observations suggest that despite high accuracy on general benchmarks, true multimodal mathematical reasoning remains an open challenge \citep{zhou2025perception,qian2025comprehensive}.

\subsection{Context Engineering}
Agentic Context Engineering (ACE) has emerged as a paradigm shift in adapting Large Language Models (LLMs) to complex tasks \citep{yao2025agenticmllm}. Popularized by Karpathy~\citep{karpathy2025context}, context engineering treats the LLM as a processor while its context window functions as working memory~\citep{langchain2025}, systematically managing information through \textit{Write}, \textit{Select}, \textit{Compress}, and \textit{Isolate} operations~\citep{anthropic2025context}. The evolution of prompting techniques has progressed from Chain-of-Thought (CoT)~\citep{wei2022chain} to ReAct~\citep{yao2023react} and Reflexion~\citep{shinn2023reflexion}, with frameworks such as LangChain~\citep{langchain2025}, AutoGPT~\citep{autogpt2025}, and the Model Context Protocol (MCP)~\citep{mcp2025} operationalizing these principles \citep{xu2025lvmagentic,ao2025emac}.

Compared to post-training methods such as Supervised Fine-Tuning (SFT) and Reinforcement Learning from Human Feedback (RLHF), ACE offers compelling advantages~\citep{arxiv2025sftlora,firoozi2025foundation}: (1) \textbf{efficiency}---no weight updates required; (2) \textbf{flexibility}---new knowledge can be injected immediately through context modification; (3) \textbf{interpretability}---each decision step can be traced through explicit context inspection; and (4) \textbf{controllability}---modifications take effect instantly without catastrophic forgetting \citep{kawaharazuka2025vla,khan2025robotics}. Nevertheless, SFT and parameter-efficient methods like LoRA may outperform pure in-context learning on specific planning tasks~\citep{arxiv2025sftlora}.

Despite rapid progress, multimodal ACE remains nascent with significant research gaps~\citep{multimodal2026survey,sui2025grounding}. Key challenges include: (1) \textbf{cross-modal alignment}---efficiently integrating heterogeneous data streams while maintaining spatiotemporal coherence \citep{oliveira2025physics}; (2) \textbf{physics constraint gap}---VLMs often fail to generate execution strategies respecting physical dynamics~\citep{vla2025challenges,zhou2025physvlm,guo2025phygrasp}; (3) \textbf{evaluation benchmarks}---existing text-centric metrics inadequately capture multimodal decision rationality \citep{elnoor2025gronav}; and (4) \textbf{inference cost}---multimodal tokens consume substantial computational resources, making adaptive token pruning an active research frontier.

\section{Diagnosing the Visual Evidence Bottleneck}
\label{sec:bottleneck}

\begin{figure}[tbp]
  \centering
  \includegraphics[width = 0.95\textwidth]{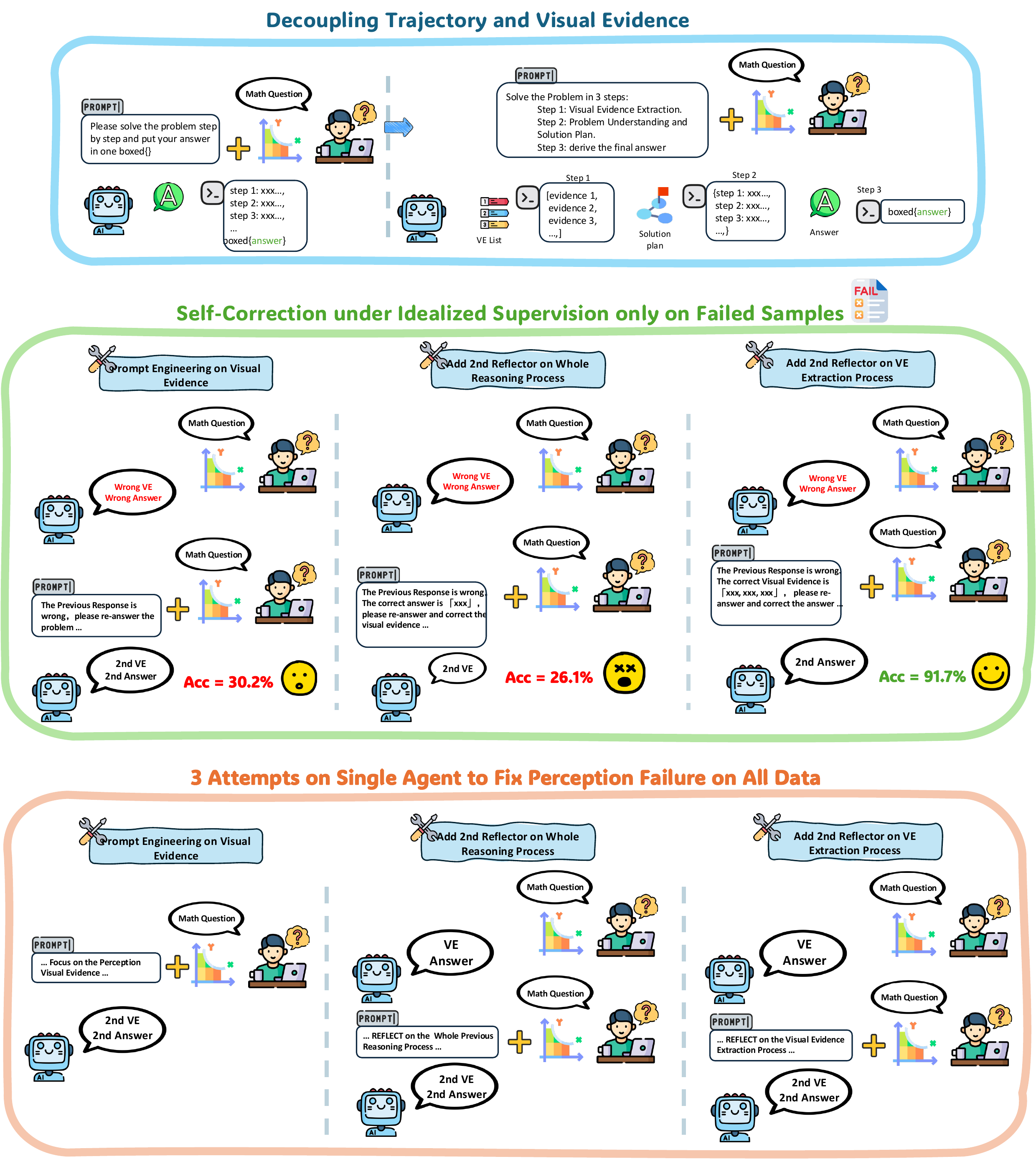}
  \caption{Diagnosing the Visual Evidence Bottleneck. 
  }
  \label{fig:diagnose}
\end{figure}


In this section, we analyze failure modes of state-of-the-art visual mathematical reasoning models and examine whether they can be resolved within a single-model setting. As shown in Figure~\ref{fig:diagnose},
by decoupling the process into visual evidence extraction, trajectory planning, and final answer using Gemini-2.5 Pro and GPT-5, we show that errors primarily originate from incorrect visual evidence rather than flawed reasoning. 
Then, self-correction experiments reveal that models struggle to recover correct visual evidence—even when given the correct answer—while supplying corrected visual evidence reliably fixes wrong answers. 
Moreover, prompt engineering and multi-round self-reflection without external guidance fail to improve visual evidence accuracy and may destabilize correct predictions. 
These findings identify visual evidence extraction as the dominant bottleneck and motivate approaches beyond isolated single-model reflection.

\subsection{Decoupling Trajectory and Visual Evidence}

\begin{table}[tbp]
  \caption{Decoupling Trajectory and Visual Evidence.}
  \label{tab: decouple}
  \centering
    \begin{tabular}{cccccc}
    \toprule
\multicolumn{2}{c}{}                                    & \multicolumn{2}{c}{Gemini 25 pro} & \multicolumn{2}{c}{GPT-5}                              \\ \cline{3-6} 
\multicolumn{2}{c}{\multirow{-2}{*}{ All 304 Samples }} & {SP steps }  & { VE list }  & {SP steps } & { VE list } \\ \hline
                                        & Total         & \multicolumn{2}{c}{208 (68.4\%)}           & \multicolumn{2}{c}{210 (69.1\%)}                                \\
                                        & \ding{51}       &\cellcolor{green!25} 205 (98.6\%)           & 170 (81.7\%)             &\cellcolor{green!25} 201 (95.7\%)           & 170   (80.9\%)                                 \\
\multirow{-3}{*}{Success }      & \ding{56}     & 3 (1.4\%)             & 38 (18.3\%)               & 9 (4.3\%)             & 40 (19.1\%)                                     \\ \hline
                                        & Total         & \multicolumn{2}{c}{96 (31.6\%)}            & \multicolumn{2}{c}{94 (30.9\%)}                                 \\
                                        & \ding{51}       &\cellcolor{green!25} 67 (69.8\%)            &\cellcolor{red!25} 12 (12.5\%)                &\cellcolor{green!25} 62 (66.0\%)            &\cellcolor{red!25} 16 (17.0\%)                                     \\
\multirow{-3}{*}{Failed }       & \ding{55}     & 29 (30.2\%)             & 84 (87.5\%)               & 32 (34.0\%)            & 78 (83.0\%)                                     \\ \hline
\multicolumn{2}{c}{Success Total}                       &\cellcolor{green!25} 272 (89.5\%)            & 182  (59.8\%)             &\cellcolor{green!25} 272 (89.5\%)           & 182 (59.8\%)                                    \\ 
    \bottomrule
    \end{tabular}
\end{table}

In this section, we explicitly decouple the model’s problem-solving process into visual perception and reasoning, and analyze their respective contributions to final answer correctness.  
As shown in the top blue  part of Figure~\ref{fig:diagnose}, given a visual math problem, the model is required to proceed in three steps: 
(1) extract visual evidence (VE) that is relevant to solving the problem; 
(2) plan a reasoning trajectory, describing the logical steps needed to reach a solution; 
and (3) instantiate the planned trajectory by grounding it in the extracted visual evidence to derive the final answer. 
This decomposition allows us to independently evaluate perception and reasoning, and to examine where failures primarily occur.

Across both Gemini-2.5 Pro and GPT-5, we observe highly consistent performance trends. 
As shown in Table~\ref{tab: decouple}, while the two models achieve comparable overall results, a clear accuracy hierarchy emerges: trajectory accuracy is the highest (close to 90\%), followed by answer accuracy (around 70\%), and finally visual evidence accuracy (around 60\%). 
When focusing on correctly answered cases, trajectory accuracy exceeds 95\% and VE accuracy remains above 80\%. In contrast, for incorrectly answered cases, trajectory accuracy remains relatively high (above 67\%), whereas VE accuracy drops sharply to approximately 12\%. 
These statistics indicate that reasoning plans are often correct even when final answers fail, and that perception errors dominate failure cases.

These findings lead to several important conclusions. 
First, a model’s solution process can be naturally decomposed—much like human problem solving—into perceived visual facts and abstract reasoning steps. 
Second, upon receiving a problem, current SOTA models already possess a strong ability to derive correct reasoning trajectories, achieving over 90\% accuracy in planning solution steps. 
Third, models significantly struggle with accurate visual evidence perception, and incorrect answers are overwhelmingly associated with severely degraded VE accuracy. Consequently, correcting failure cases fundamentally requires improving visual evidence accuracy; without reliable perception, even strong reasoning capabilities cannot yield correct answers.

\subsection{Self-Correction under Idealized Supervision}

\begin{table}[tbp]
  \caption{Self-Correction under Idealized Supervision}
  \label{tab: self-correction}
  \centering
    \begin{tabular}{ccccc}
    \toprule
\multirow{2}{*}{Add Info} & \multicolumn{2}{c}{Gemini 2.5 pro} & \multicolumn{2}{c}{GPT 5} \\ \cline{2-5} 
                          & VE          & Answer         & VE     & Answer     \\ \hline
GT judge                  & 22/96          & 29/96             & 25/94     & 31/94         \\
GT judge + GT answer          &\cellcolor{red!25} 22/84          & -                 &\cellcolor{red!25} 22/78     & -             \\
GT VE                     & -              & \cellcolor{green!25}88/96             & -         & \cellcolor{green!25}89/94         \\ 
    
    \bottomrule
    \end{tabular}
\end{table}


In this subsection, we evaluate the feasibility of model self-correction under idealized conditions.
Experiments are conducted on the error subset of the mini benchmark, where models fail in the first round.
To simulate external supervision, we retain the full first-round interaction history and require the model to re-answer with additional information provided in the prompt. 

As shown in the middle green  part of Figure~\ref{fig:diagnose}, we consider three progressively stronger forms of supervision: 
(1) providing ground-truth judgment, explicitly informing the model that its previous answer was incorrect; 
(2) in cases where visual evidence is incorrect, providing both ground-truth judgment and the correct final answer, and asking the model to revise its visual evidence; 
and (3) directly providing corrected visual evidence, while requiring the model to re-answer the question.

As shown in Table~\ref{tab: self-correction}, our results reveal clear limitations in the model’s ability to self-correct. 
Under the most basic supervision—explicitly stating that the previous answer is wrong—the model achieves only around 30\% second-round accuracy, which serves as an approximate upper bound for self-reflection-based correction. 
When additionally providing the correct final answer, the model’s ability to retrospectively correct its visual evidence remains limited, with VE correction rates below 30\%. 
In contrast, when erroneous visual evidence is minimally corrected and fed back to the model, answer accuracy improves dramatically, reaching 88.5\% and 84.6\%, respectively. 
Notably, even after being informed of its mistake, the model frequently persists in its original answer, exhibiting a form of overconfidence or rigidity; we provide a more detailed analysis of this behavior in the appendix.

These findings lead to three key conclusions. 
First, current models possess a limited but non-negligible capacity for second-round reflection, achieving up to 30\% correction accuracy when explicitly told they are wrong. 
Second, the primary difficulty lies in visual evidence perception: models lack the ability to effectively backtrack and reconstruct correct visual evidence, even when the correct answer is provided. 
Third, extracting complete and accurate visual evidence is the true bottleneck—while ground-truth visual evidence reliably yields correct answers, the inverse does not hold, as correct answers rarely enable the recovery of correct visual evidence.

\subsection{Limits of Single-Model Prompting and Reflection}

\begin{table}[tbp]
  \caption{Limits of Single-Model Prompting and Reflection}
  \label{tab: limit single agent reflect}
  \centering
    \begin{tabular}{cllllll}
    \toprule
    \multirow{2}{*}{Reflect Mode} & \multicolumn{2}{c}{Gemini2.5 pro} & \multicolumn{2}{c}{GPT 5} \\ \cline{2-5} 
                                  & VE            & Answer           & VE        & Answer       \\ \hline
    Orignal                       & 182             & 208              & 186        & 210          \\
    1st turn prompt engineering     & 181$_{\textcolor{red}{-1}}$          & 212$_{\textcolor{green}{+4}}$            & 190$_{\textcolor{green}{+4}}$      & 207$_{\textcolor{red}{-3}}$           \\
    2nd turn reflect on answer      & 189$_{\textcolor{green}{+7}}$          & 204$_{\textcolor{red}{-4}}$            &169$_{\textcolor{red}{-17}}$     & 202$_{\textcolor{red}{-8}}$        \\
    2nd turn reflect on VE          & 174$_{\textcolor{red}{-8}}$          & 200$_{\textcolor{red}{-8}}$            & 180$_{\textcolor{red}{-6 }}$     & 210$_{\textcolor{gray}{+0}}$          \\
    \bottomrule
    \end{tabular}

\end{table}


In this subsection, we examine whether a single model can independently resolve visual evidence perception errors without any external supervision. 
Specifically, we explore several commonly adopted prompt-based strategies that aim to improve perception through better instruction or self-reflection. 
As shown in the bottom orange part of Figure~\ref{fig:diagnose},
First, we emphasize the importance of visual evidence in the initial prompt, explicitly warning the model that incorrect VE extraction will lead to incorrect answers and instructing it to carefully verify visual evidence before proceeding with reasoning. 
Second, we introduce a second-round reflection stage where the model is asked to reflect on its entire first-round reasoning process. 
Third, we further narrow the focus of reflection by explicitly instructing the model to revisit and revise its visual evidence extraction before re-answering.

Despite these efforts, we observe no substantial improvement in either obvious visual evidence accuracy or final answer accuracy. 
As shown in Table~\ref{tab: limit single agent reflect},
emphasizing visual evidence in the prompt—whether in the first round or during second-round reflection—fails to meaningfully improve VE correctness. 
Moreover, when no new information is introduced, applying second-round reflection across all samples yields minimal gains, regardless of whether the reflection targets the full reasoning process or explicitly focuses on visual evidence. 
These results suggest that prompt-level emphasis alone is insufficient to overcome perception errors.

We conclude that simple prompt adjustments within a single-model setting are inadequate for correcting visual perception limitations. 
Multi-round self-reflection without external posterior guidance does not reliably converge: the model struggles to fix incorrect cases and may even degrade previously correct answers by introducing new errors. 
As detailed in the appendix, this instability further underscores the inherent difficulty of perception correction under a single-model paradigm and motivates the need for alternative strategies beyond isolated self-reflection.

\subsection{Summary and Motivation for Multi-Agent Visual Evidence Modeling}
We summarize the above analyses as follows. 
First, by explicitly decoupling the problem-solving process into visual evidence extraction, solution trajectory planning, and final reasoning, we demonstrate that failures in visual mathematical reasoning primarily stem from errors in visual evidence perception rather than deficiencies in reasoning ability.
While models are generally capable of deriving correct solution trajectories, inaccurate or incomplete visual evidence fundamentally undermines the final answer.

Second, we investigate the feasibility of correcting incorrect cases through model self-correction with external posterior supervision. 
Under idealized conditions—where the model is explicitly informed of its mistake—the correction rate remains limited to around 30\%. Moreover, even when provided with the ground-truth final answer, the model exhibits weak capability in retrospectively reconstructing correct visual evidence. 
In contrast, supplying corrected visual evidence reliably leads to correct answers, establishing that correct visual evidence is a prerequisite for correct answers, while the reverse does not hold.

Finally, we remove all external posterior information and evaluate whether a single model can independently resolve visual evidence errors. 
Our results show that neither emphasizing visual evidence in the prompt nor applying multi-round self-reflection effectively mitigates perception errors. Without explicit external guidance, reflection not only fails to correct incorrect cases but may also destabilize correct ones, causing previously correct answers to degrade. 
These findings indicate that single-model prompting and self-reflection are insufficient to overcome the inherent limitations of visual perception.

To address these challenges and transcend the limitations of a single agent, we propose to introduce \textbf{multiple agents} to collaboratively and dynamically maintain the core bottleneck—namely, the visual evidence list. 
Different agents naturally offer diverse and sometimes conflicting interpretations of visual evidence. 
Compared to prompt-level emphasis, such inter-agent disagreement provides a more explicit and concrete signal that encourages models to re-examine their own perception. 
Furthermore, the patterns of agreement and conflict among multiple experts partially emulate reliable posterior supervision, guiding second-round reflection while reducing the risk of degrading correct answers. 
This motivation directly leads to our core design principles, which propose a multi-agent context engineering framework on visual perceptions.

\section{Method}
\label{sec: method}

\begin{figure}[tbp]
  \centering
  \includegraphics[width = 1\textwidth]{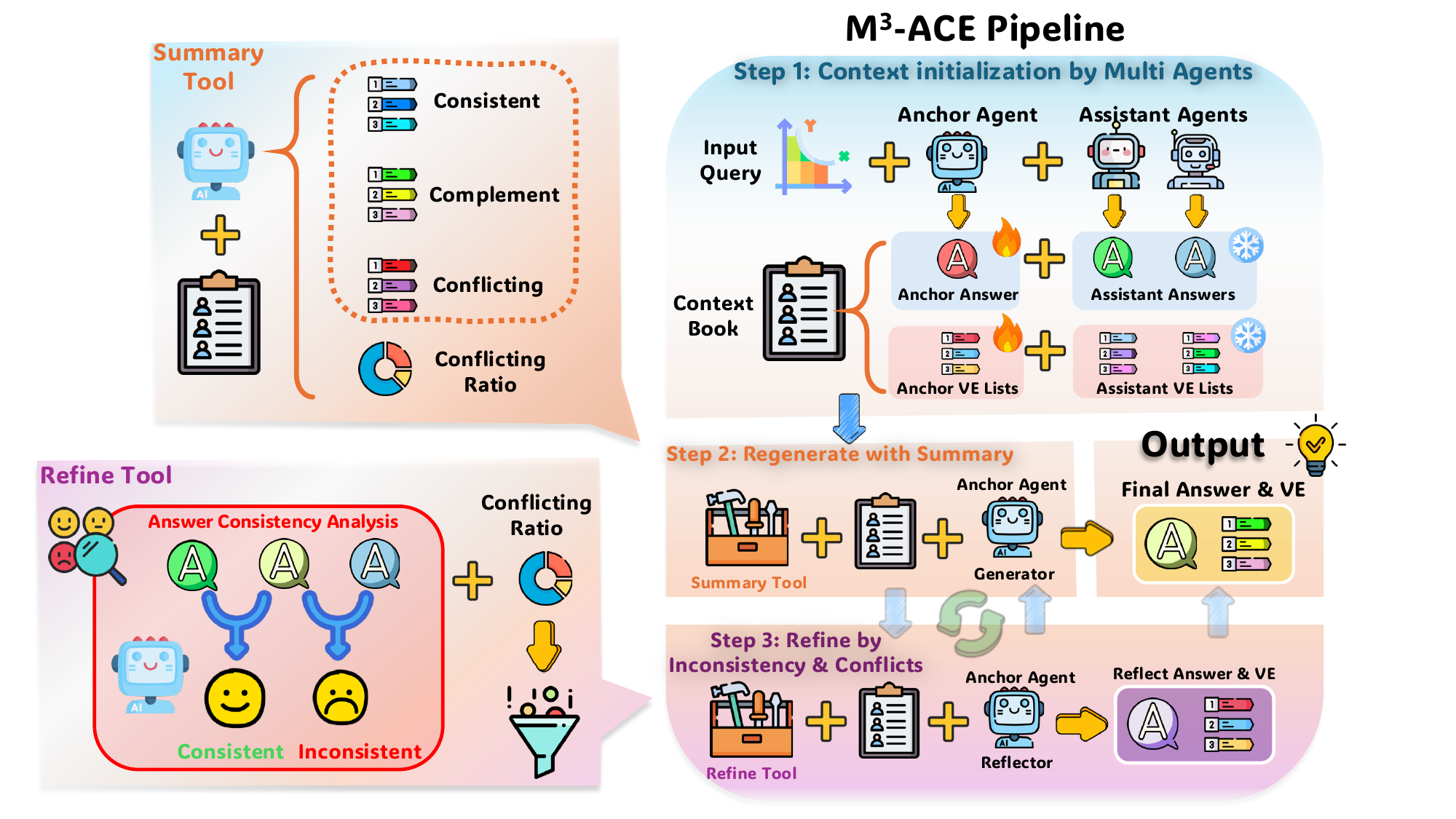}
  \caption{M$^3$-ACE Pipeline(right part) with Two  Auxiliary Tools(left part).
  }
  \label{fig:pipeline}
\end{figure}

In this section, we introduce M3-Agent, a multi-agent framework designed to progressively address the visual evidence bottleneck identified in previous sections. Our system is built upon three core design principles:
\begin{enumerate}
    \item \textbf{Decoupling Principle.} To directly target the bottleneck, we explicitly separate the visual evidence list from the final answer in the original response. For the anchor agent, both its visual evidence list and answer are treated as the primary components of an iteratively updated context. 
    \item \textbf{Complementary Information Principle.} To mitigate the anchor agent’s tendency to over-trust its own initial perception, we introduce heterogeneous assistant agents whose responses serve as external information sources. By exposing the anchor agent to diverse and potentially conflicting visual evidence, we encourage reconsideration during refinement and reduce confirmation bias in self-reflection.
    \item \textbf{Filtering Principle.} To improve iteration efficiency and reduce redundant token consumption, we selectively focus on difficult or highly disputed samples. Easy or high-consensus cases are filtered out early, and iterative refinement is applied primarily to samples with substantial disagreement or perceptual inconsistency.
\end{enumerate}

\subsection{M$^3$-ACE Pipeline}
Based on these principles, we design the M$^3$-ACE pipeline, which consists of three main steps supported by two auxiliary tools.

\textbf{Step 1: Multi-Agent Initialization.}
We first invite the anchor agent and assistant agents to independently respond to the problem. Each response is explicitly structured into two parts: a visual evidence list and a final answer. These components collectively form a shared context repository for solving the problem. The anchor agent’s context is treated as an iteratively refinable component, while assistant agents’ contexts are frozen as external reference signals and are not updated in later iterations.

\textbf{Step 2: Context Summarization and Regeneration.}
Next, we invoke a summary tool to aggregate and categorize the visual evidence lists from multiple agents into three groups: consistent, complementary, and conflicting. Based on this summarized context, the anchor agent regenerates an updated visual evidence list and corresponding answer. This step enables structured incorporation of inter-agent agreement and disagreement into the refinement process.

\textbf{Step 3: Refinement and Filtering.}
We then apply a refine tool to analyze (i) the consistency between the new and previous answers and (ii) the proportion of inconsistent visual evidence within the updated context. Samples whose new answers diverge significantly from prior consensus and contain a high ratio of conflicting visual evidence are rejected and sent back to Step 2 for further refinement. The remaining samples are selected as finalized predictions.

The full pipeline is detailed in Algorithm~\ref{alg:m3agent}. The refinement loop (Steps 2–3) continues until the ratio of selected samples reaches the predefined threshold, at which point the algorithm terminates.

\begin{algorithm}[tbp]
\caption{M$^3$-ACE: Multi-Agent Context Engineering}
\label{alg:m3agent}
\begin{algorithmic}[1]
\REQUIRE Problem set $\mathcal{P}=\{(Q_j,I_j)\}_{j=1}^{M}$; 
Anchor agent $A_0$; 
Assistant agents $\{A_1,\dots,A_n\}$; 
Selection ratio threshold $\tau$
\ENSURE Final answers $\{\hat{y}_j\}$

\STATE \textbf{// Step 1: Multi-Agent Initialization}
\FOR{each problem $(Q_j, I_j) \in \mathcal{P}$}
    \FOR{each agent $A_i \in \{A_0, A_1, ..., A_n\}$}
        \STATE $(VE_{i,j}^{(0)}, y_{i,j}^{(0)}) \leftarrow A_i(Q_j, I_j)$
    \ENDFOR
    \STATE Freeze assistant context $\{(VE_{i,j}^{(0)}, y_{i,j}^{(0)})\}_{i=1}^{n}$
    \STATE Initialize anchor context 
    $C_{j}^{(0)} \leftarrow (VE_{0,j}^{(0)}, y_{0,j}^{(0)})$
\ENDFOR

\STATE $\mathcal{U} \leftarrow \mathcal{P}$ \hfill // Unresolved problems
\STATE $\mathcal{S} \leftarrow \emptyset$ \hfill // Selected problems
\STATE $k \leftarrow 0$

\REPEAT
    \FOR{each problem $j \in \mathcal{U}$}

        \STATE \textbf{// Step 2: Context Summarization}
        \STATE $(VE^{cons}, VE^{comp}, VE^{conf}) 
        \leftarrow \textsc{SummaryTool}(C_{j}^{(k)}, \{VE_{i,j}^{(0)}\}_{i=1}^{n})$

        \STATE $(VE_{0,j}^{(k+1)}, y_{0,j}^{(k+1)}) 
        \leftarrow A_0(Q_j, I_j \mid VE^{cons}, VE^{comp}, VE^{conf})$

        \STATE \textbf{// Step 3: Refinement and Filtering}
        \STATE decision $\leftarrow$ 
        \textsc{RefineTool}$(y_{0,j}^{(k+1)}, \{y_{i,j}^{(0)}\}_{i=1}^{n}, VE_{0,j}^{(k+1)})$

        \IF{decision == Select}
            \STATE $\mathcal{S} \leftarrow \mathcal{S} \cup \{j\}$
            \STATE $\hat{y}_j \leftarrow y_{0,j}^{(k+1)}$
        \ELSE
            \STATE Keep $j$ in $\mathcal{U}$ for next iteration
        \ENDIF

    \ENDFOR

    \STATE $\mathcal{U} \leftarrow \mathcal{P} \setminus \mathcal{S}$
    \STATE $k \leftarrow k + 1$

\UNTIL{$|\mathcal{S}| / |\mathcal{P}| \geq \tau$}

\RETURN $\{\hat{y}_j\}_{j \in \mathcal{S}}$
\end{algorithmic}
\end{algorithm}

\subsection{Two  Auxiliary Tools}
To support the dynamic maintenance of the multi-agent context, we introduce two external auxiliary modules: a Summary Tool and a Refine Tool. Both tools are driven by lightweight agents and combined with explicitly designed hard logical rules to ensure stable summarization and filtering. Unlike the main anchor agent, these tools are not responsible for solving the task directly; instead, they structure and regulate the iterative refinement process.

\textbf{Summary Tool.}
The Summary Tool is responsible for aggregating and structuring visual evidence across agents. 
It takes as input the current anchor agent context and the frozen assistant agent contexts. 
Specifically, given the visual evidence list from the anchor agent, the tool compares each item against the visual evidence lists provided by assistant agents. 
Based on this comparison, visual evidence items are categorized into three groups relative to the anchor agent:
\begin{enumerate}
    \item Consistent VE: Evidence supported by one or more assistant agents.
    \item Complementary VE: Evidence provided by assistant agents but absent from the anchor agent’s list.
    \item Conflicting VE: Evidence that contradicts the anchor agent’s extracted visual facts.
\end{enumerate}
In addition to this tripartite classification, the tool computes a conflict ratio, defined as the proportion of conflicting visual evidence items relative to the total number of visual evidence items in the anchor context. 
The Summary Tool outputs the three categorized VE groups along with the conflict ratio, which serves as a quantitative signal of perceptual instability for downstream refinement.

\textbf{Refine Tool.}
The Refine Tool determines whether a problem instance should be accepted or further iterated. It operates in two stages.

First, a lightweight agent evaluates answer consistency between the anchor agent and assistant agents. If the number of assistant answers agreeing with the anchor answer is greater than or equal to half of the total number of assistant agents, the response is considered high-consensus.

Second, this answer consistency signal is combined with the previously computed conflict ratio. If a sample exhibits both (i) a high conflict ratio (greater than 0.2) and (ii) is not classified as high-consensus, it is rejected and returned to the next iteration for further refinement. All remaining samples are selected and retained as finalized outputs.

Through this combination of lightweight semantic analysis and hard logical filtering, the Summary and Refine tools jointly enable stable and efficient iterative context maintenance, preventing unnecessary re-computation on easy samples while focusing refinement efforts on perceptually ambiguous and disputed cases.

\section{Main Result}
\label{sec:result}
\subsection{Results on the MathVision Benchmark}

In this section, we evaluate the effectiveness of \textbf{M3-Agent} on the MathVision benchmark. 
Our experiments aim to verify two key design motivations: 
(1) introducing external information from multiple agents, and 
(2) decoupling visual evidence (VE) from answers to enable iterative context engineering.

\paragraph{Experimental Setup.}
We construct two groups of agent ensembles with different capability levels: a \textbf{SOTA ensemble} and a \textbf{sub-SOTA ensemble}. 
The SOTA ensemble consists of four state-of-the-art multimodal large language models: 
Gemini-3 Pro, Gemini-2.5 Pro, GPT-5, and Claude-4.5 Sonnet. 
Among them, Gemini-3 Pro achieves the best single-model performance on MathVision (over 80\% accuracy), 
while Claude-4.5 Sonnet performs the worst (around 60\%). 
To form the sub-SOTA ensemble, we remove both Gemini-3 Pro and Claude-4.5 Sonnet, 
retaining the remaining models.

During inference, we alternately designate each model in the ensemble as the \textit{anchor agent}, 
while the remaining models serve as \textit{assistant agents}. 
The full results are reported in Table~\ref{tab: sota mathvision} and Table~\ref{tab: sub-sota mathvision}.

\paragraph{Impact of Multi-Agent Information.}
Introducing external viewpoints from multiple agents consistently improves performance. 
As shown in the \textit{1st Regenerate} column of Table~\ref{tab: sota mathvision} and Table~\ref{tab: sub-sota mathvision}, all anchor models achieve higher accuracy compared with their direct single-agent inference results.

Within the SOTA ensemble, Gemini-3 Pro improves by 3.1 points, 
Gemini-2.5 Pro by 7.2 points, 
GPT-5 by 9.3 points, 
and Claude-4.5 Sonnet by 14.8 points. 
Similar improvements are observed in the sub-SOTA ensemble for Gemini-2.5 Pro, GPT-5, and Claude-4.5 Sonnet.

These results indicate that incorporating perspectives from other agents provides valuable complementary information for solving visual math problems. 
For relatively weaker models (e.g., Claude-4.5 Sonnet), the inclusion of stronger agents introduces correct evidence that helps correct perceptual errors, resulting in large accuracy gains. 
Meanwhile, stronger models (e.g., Gemini-3 Pro in the SOTA ensemble and Gemini-2.5 Pro in the sub-SOTA ensemble) are not negatively affected by weaker agents. 
Instead, they selectively absorb useful information while ignoring incorrect evidence, achieving additional performance improvements. 
Overall, the consistent gains demonstrate the effectiveness of the multi-agent design in M3-Agent.

\paragraph{Benefits of Multi-Round Reflection.}
Beyond the multi-agent setup, iterative reasoning further improves performance.

During the \textit{refine} stage, the refine tool selects answers based on both VE lists and answer consistency across agents. 
As shown in the \textit{1st Refine} column of Table~\ref{tab: sota mathvision}, the selected subset achieves an accuracy close to 90\%, which is approximately three points higher than the overall average accuracy. 
In contrast, rejected samples have an accuracy more than 30 points lower than the average. 
This indicates that the refine stage effectively filters out a small number of low-quality predictions.

Importantly, only about one-tenth of the samples enter the iterative correction process, 
which significantly improves computational efficiency for the entire pipeline.

During the subsequent \textit{reflect} stage, agents iteratively revise their predictions on these difficult samples. 
Substantial improvements are observed: Gemini-3 Pro improves by 10.8 points, 
Gemini-2.5 Pro by 9.4 points, 
GPT-5 by 8.5 points, 
and Claude-4.5 Sonnet by 1.0 point. 
These results show that the refine–reflect mechanism effectively resolves hard cases through targeted multi-round reasoning.

\paragraph{Performance Gains Across Model Capability Levels.}
An interesting pattern emerges when comparing models with different capability levels. 
Stronger models benefit less from multi-agent information but gain more from multi-round reflection, 
while weaker models show the opposite trend.

For example, in the \textit{1st Regenerate} stage, Gemini-3 Pro gains only 3.1 points, 
which is much smaller than the 14.8-point improvement achieved by Claude-4.5 Sonnet. 
However, during the reflection stage, Gemini-3 Pro improves by 10.5 points, 
significantly larger than the 1.0-point gain of Claude-4.5 Sonnet.

This observation aligns with intuitive expectations. 
For high-capability agents, external opinions from weaker models have limited influence, 
but they possess strong self-correction ability when explicitly prompted to reconsider their reasoning. 
Conversely, weaker models benefit greatly from correct external evidence provided by stronger agents, 
but often lack the perceptual capability required for effective self-correction.

These findings suggest that strong agents excel at understanding and correcting perceptual details, 
while weaker agents rely more heavily on external guidance. 
The combined refine–reflect design of M$^3$-ACE effectively leverages both mechanisms, 
allowing the framework to accommodate agents of different capability levels and demonstrating strong generalization ability.

\begin{table}[tbp]
  \caption{
Results of the SOTA Ensemble.
Subscripts without uparrows($\uparrow$) denote the number of samples. The green numbers indicate the number of samples filtered as Select, while the red numbers indicate the number of samples filtered as Reject. The selected samples are merged into the final answer pool, whereas the rejected samples are regenerated and proceed to the next round.
Subscripts with arrows represent the accuracy (ACC) improvement compared to the previous stage. The gray arrow subscripts indicate the ACC gain of the first-round “Regenerate with Summary” compared to CoT inference. The blue arrow subscripts denote the ACC improvement obtained after the 1st Reflect step on the rejected samples, relative to the answers from the first-round “Regenerate with Summary.” The purple arrow subscripts represent the overall ACC improvement on the entire dataset, from the first-round “Regenerate with Summary” to after the 1st Reflect stage.
}
  \label{tab: sota mathvision}

\begin{tabular}{ccccccc}
\toprule
\multirow{2}{*}{\makecell{Anchor \\ Agent}} & \multirow{2}{*}{CoT Infer} & \multirow{2}{*}{\makecell{1st Regenerate \\ with Summary}} & \multicolumn{2}{c}{1st Refine} & \multicolumn{2}{c}{1st  Reflect} \\
                  &                            &           & Select         & Reject        & Reject        & All       \\ \midrule
Gemini 3 pro      & 85.0                       & 88.1 {\color{gray}\scriptsize3.1$\uparrow$}      & 92.7$_{\color{green}2741}$  & 46.1$_{\color{red}299}$  & 56.9 {\color{blue}\scriptsize10.8$\uparrow$}           & 89.1 {\color{purple}\scriptsize1.0$\uparrow$}             \\
Gemini 2.5 pro    & 73.3                       & 80.5 {\color{gray}\scriptsize7.2$\uparrow$}      & 84.4$_{\color{green}2812}$  & 32.2$_{\color{red}228}$  & 41.6 {\color{blue}\scriptsize9.4$\uparrow$}         & 81.2 {\color{purple}\scriptsize0.7$\uparrow$}             \\
GPT 5             & 72.0                       & 81.3 {\color{gray}\scriptsize9.3$\uparrow$}      & 85.5$_{\color{green}2709}$  & 47.1$_{\color{red}331}$  & 55.6 {\color{blue}\scriptsize8.5$\uparrow$}         & 82.2 {\color{purple}\scriptsize0.9$\uparrow$}              \\
Claude-45         & 61.2                       & 76.0 {\color{gray}\scriptsize14.8$\uparrow$}      & 79.2$_{\color{green}2250}$  & 66.7$_{\color{red}790}$  & 68.7 {\color{blue}\scriptsize1.0$\uparrow$}         & 76.5 {\color{purple}\scriptsize0.5$\uparrow$}              \\  

 \bottomrule
\end{tabular}
\end{table}

\begin{table}[tbp]
  \caption{Results of the sub-SOTA Ensemble.}
  \label{tab: sub-sota mathvision}
    \centering
\begin{tabular}{cccc}
\toprule
\multirow{2}{*}{\makecell{Anchor \\ Agent}} & \multirow{2}{*}{CoT Infer} & \multirow{2}{*}{\makecell{1st Regenerate \\ with Summary}} \\
                  &                            &                                    \\ \hline
Gemini 2.5 pro    & 73.3                       & 77.6                                                 \\
GPT 5             & 72.0                       & 78.5                                                 \\
Claude-45         & 61.2                       & 71.2                                                \\ 

 \bottomrule
\end{tabular}
\end{table}



\subsection{Results on Other Math Vision Benchmarks}

We further evaluate M$^3$-ACE on additional visual mathematical reasoning benchmarks, including MathVista and MathVerse. 
Across these datasets, the proposed framework consistently achieves notable performance improvements over single-model inference. 
These results demonstrate that the benefits of multi-agent context engineering and VE–answer decoupling generalize beyond the MathVision benchmark. 
Detailed experimental results are provided in the Appendix.

\section{Ablation Studies}

In this section, we present additional ablation studies to further validate our previous findings and provide deeper analysis of the proposed framework. 
In particular, we investigate the following questions: 
(1) whether the improvement is indeed related to the correction of visual perception errors; 
(2) whether the key bottleneck lies in visual evidence rather than the final answer aggregation; 
(3) how weaker agents influence stronger agents when introduced into the multi-agent framework; and 
(4) the stability and convergence behavior of the reflection iterations.

\paragraph{Q1: Does performance improvement correlate with corrected visual perception?}
To verify whether the observed gains stem from improved visual perception, we perform additional sampling-based evaluation on the generated VE lists. 
The results show that accuracy improvements strongly correlate with corrected visual evidence, confirming our earlier hypothesis that perception errors constitute the primary bottleneck. 
More detailed analysis is provided in the Appendix.

\paragraph{Q2: Context engineering on VE vs. answer-level multi-agent aggregation.}
We further compare our approach with alternative strategies that directly aggregate final answers across multiple agents and rounds. 
The results show that maintaining the intermediate perception process—specifically the VE list—as the shared context leads to significantly better performance than simply aggregating answers. 
Although this design introduces a slightly more complex pipeline, it directly targets the core bottleneck and yields more reliable improvements. 
Detailed results are reported in the Appendix.

\paragraph{Q3: Influence of weaker agents on stronger agents.}
We conduct additional experiments by grouping agents according to their standalone reasoning performance and evaluating combinations of stronger and weaker models. 
Under the guidance of the proposed summary tool, stronger agents consistently benefit from complementary observations provided by both peer-level and weaker agents. 
From a posterior analysis perspective, we further visualize the upset patterns among agent predictions, providing insight into the phenomenon where weaker agents occasionally supply crucial missing evidence that helps stronger agents correct perception errors. 
Further details are included in the Appendix.

\paragraph{Q4: Stability of the reflection process.}
Finally, we analyze the stability and convergence of the reflection iterations under different anchor agents and refine conditions. 
The experiments reveal that the refine tool plays a critical role in stabilizing the reflection process by filtering unreliable cases and preventing degeneration during iterative reasoning. 
Additional experimental results and analysis are provided in the Appendix.

\section{Conclusion}

In this work, we systematically investigate the failure modes of state-of-the-art multimodal large language models in visual mathematical reasoning. 
Through a series of controlled analyses, we identify that the primary bottleneck does not lie in the reasoning capability itself, but rather in the perception stage—specifically in the extraction of accurate \emph{visual evidence} from images. 
Our experiments show that models are generally capable of producing correct reasoning trajectories when the necessary information is available, yet they frequently fail to perceive or correctly describe key visual facts. 
Moreover, models tend to remain overly confident in these incorrect perceptions, making it difficult for conventional approaches such as prompt refinement, multi-round self-reflection, or external posterior guidance to reliably correct the errors. This combination of perception errors and persistent incorrect beliefs forms the core obstacle preventing further improvements in visual math reasoning.

To address this limitation, we propose \textbf{M$^3$-ACE}, a multi-agent, multi-turn multimodal context engineering framework designed to explicitly target the perception bottleneck. 
Instead of directly aggregating final answers, our approach decouples the reasoning process by focusing on the maintenance and refinement of a shared \emph{visual evidence list}. 
Multiple agents with diverse perception capabilities collaboratively contribute observations, allowing complementary information to emerge while exposing potential conflicts in visual interpretation. 
The system dynamically maintains a context book centered on these visual evidence lists, which serve as the foundation for subsequent reasoning and answer generation. 
To ensure efficient coordination and stable convergence across agents and iterations, we further introduce two lightweight auxiliary tools: a \textit{Summary Tool}, which aggregates and categorizes visual evidence from multiple agents into consistent, complementary, and conflicting components, and a \textit{Refine Tool}, which selectively filters unreliable samples and guides the iterative reflection process.

Extensive experiments across multiple visual mathematical reasoning benchmarks demonstrate that the proposed framework significantly improves performance compared with single-agent inference and naive multi-agent aggregation strategies. 
The results show that incorporating diverse perceptual perspectives effectively corrects missing or incorrect visual evidence, while the iterative refinement process stabilizes multi-round reasoning and prevents degeneration during reflection. 
Notably, even strong models can benefit from complementary observations provided by weaker agents, highlighting the importance of diverse perception in multimodal reasoning systems.

Overall, our findings suggest that improving visual perception and its integration into reasoning pipelines is crucial for advancing multimodal mathematical reasoning. 
By shifting the focus from answer-level aggregation to perception-level context engineering, M$^3$-ACE provides a new perspective on how multi-agent collaboration can address fundamental limitations of current models. 
We hope this work encourages further exploration of structured multi-agent reasoning frameworks and perception-aware context engineering for more reliable and robust multimodal intelligence.

\appendix

\small
\bibliography{ref}
\normalsize
\end{document}